\documentclass[10pt,twocolumn,letterpaper]{article}

\usepackage{iccv}
\usepackage{times}
\usepackage{epsfig}
\usepackage{graphicx}
\usepackage{amsmath}
\usepackage{amssymb}
\usepackage{epstopdf}
\usepackage{multirow}
\usepackage{verbatim}
\usepackage[pagebackref=true,breaklinks=true,letterpaper=true,colorlinks,bookmarks=false]{hyperref}

\iccvfinalcopy 

\def\iccvPaperID{3622} 

\ificcvfinal\fi
\begin{document}

\title{Second-order Non-local Attention Networks for Person Re-identification}

\author{
    Bryan (Ning) Xia, Yuan Gong, Yizhe Zhang, Christian Poellabauer\\
    University of Notre Dame\\
    Notre Dame, IN 46556 USA\\
    {\tt\small \{nxia, ygong1, yzhang29, cpoellab\}@nd.edu}
}

\maketitle
\thispagestyle{empty}

\begin{abstract}
Recent efforts have shown promising results for person re-identification by designing part-based architectures to allow a neural network to learn discriminative representations from semantically coherent parts. Some efforts use soft attention to reallocate distant outliers to their most similar parts, while others adjust part granularity to incorporate more distant positions for learning the relationships. Others seek to generalize part-based methods by introducing a dropout mechanism on consecutive regions of the feature map to enhance distant region relationships. However, only few prior efforts model the distant or non-local positions of the feature map directly for the person re-ID task. In this paper, we propose a novel attention mechanism to \textbf{directly} model long-range relationships via second-order feature statistics. When combined with a generalized DropBlock module, our method performs equally to or better than state-of-the-art results for mainstream person re-identification datasets, including Market1501, CUHK03, and DukeMTMC-reID.
\end{abstract}

\section{Introduction}
Person re-identification (re-ID) is an essential component of intelligent surveillance systems, which draws increasing interest from the computer vision community. It is challenging to associate multiple images captured by cameras with non-overlapping viewpoints with the same person-of-interest. Specifically, this task is challenging due to the dramatic variations with respect to illumination, occlusion, resolution, human pose, view angle, clothing, and background. The person re-ID research community has proposed various effective hand-crafted features
~\cite{55bazzani2010multiple, 57li2013learning, 58mignon2012pcca, 59pedagadi2013local, 60ma2013domain, 61das2014consistent, 62liao2015person, 21matsukawa2016hierarchical}
to address these challenges. Methods based on deep convolutional networks have also been introduced to learn discriminative features and representations that are robust to these variations, thereby pushing multiple re-ID benchmarks to a whole new level. Among these methods, several efforts~\cite{23shen2018deep, 05sun2018beyond, 07wang2018learning, 24zheng2018coarse} learn detailed features from local parts of a person's image, while others extract useful global features~\cite{43sun2017svdnet, 42zheng2018pan, 10chen2018group, 04dai2018batch}.

Recently, part-based models~\cite{05sun2018beyond, 07wang2018learning, 24zheng2018coarse} have made great progress towards learning effective part-informed representations for person re-ID, achieving very promising results. By partitioning the backbone network's feature map horizontally into multiple parts, the deep neural networks can concentrate on learning more fine-grained salient features in each individual local part. The aggregation of these features from all parts provides discriminative cues for each identity as a whole. However, these models, on one hand, suffer from one common drawback: they require relatively well-aligned body parts for the same person in order to learn salient part features. On the other hand, strict uniform partitioning of the feature map breaks within-part consistency. Several recent efforts proposed different remedies to compensate for the side effects of part partitioning, which are described below.

\begin{figure}[t]
\centering
\includegraphics[width=8.3cm]{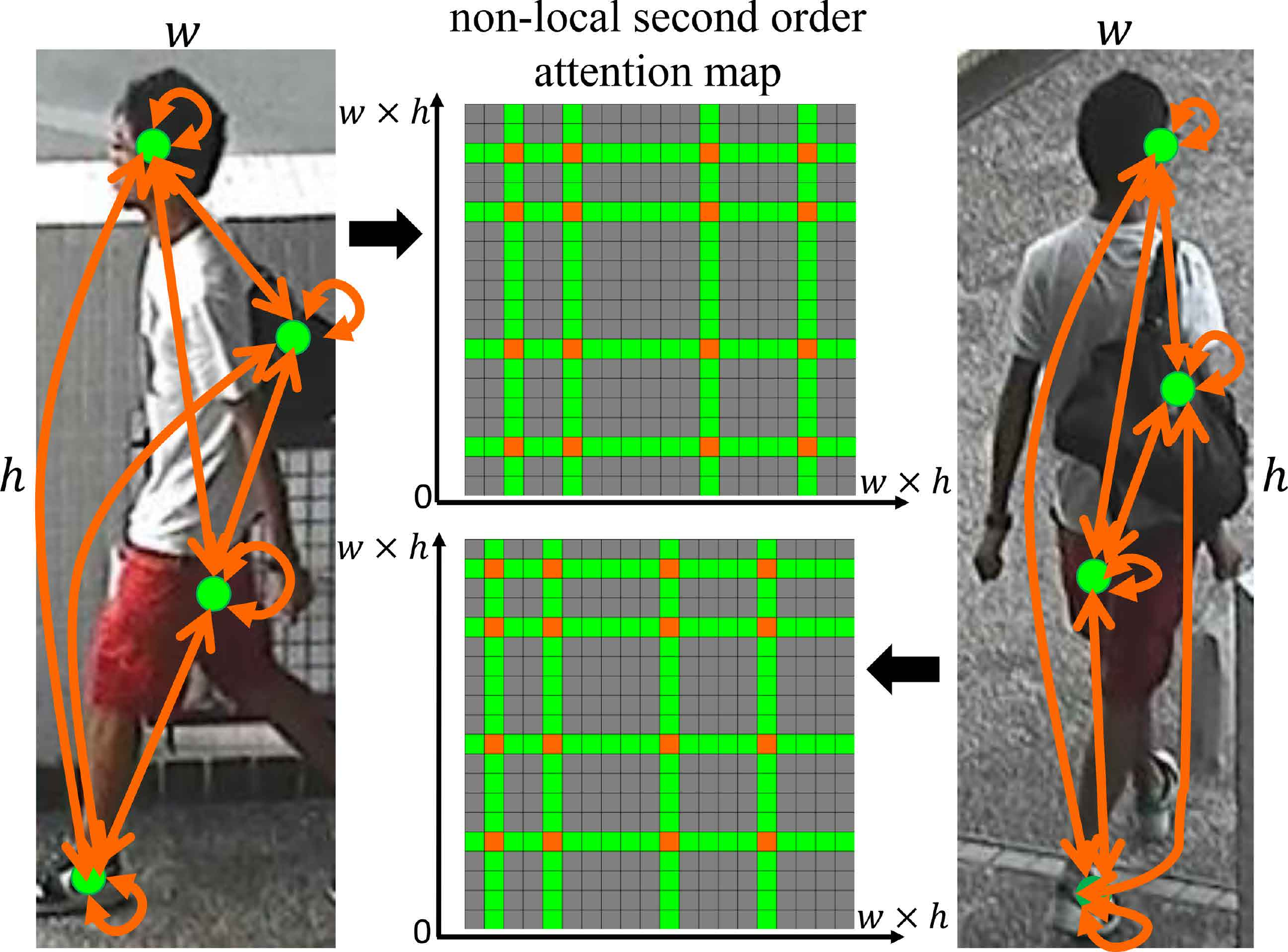}
\caption{Illustration of second-order non-local attention for person re-identification. We show images from two views of one person and illustration of the attention map. Our second-order non-local attention map allows the model to learn to encode non-local part-to-part correlations (marked in orange).}
\label{fig:figure0}
\end{figure}

When related image areas fall into other parts, Part-based Convolutional Baseline (PCB)~\cite{05sun2018beyond} addresses the misalignment by rearranging the part partition by enforcing part consistency using soft attention. Although this treatment allows for a more robust part partition, the initial rigid uniform partition of the feature map still greatly limits the representation learning capability of a deep learning model. As observed by the authors of PCB~\cite{05sun2018beyond}, when the number of parts increases, the accuracy does not increase monotonically. When the part number increases, it breaks the part coherence, making it difficult for the deep neural network to capture meaningful information from the parts, thereby harming the performance. PCB also ignores global feature learning, which captures the most salient features to represent different identities~\cite{07wang2018learning}, losing the opportunity to consider the feature map as a semantic part (distinguished from unrelated background).

Multiple Granularity Network (MGN)~\cite{07wang2018learning} improves PCB by adding a global branch to treat the whole feature map as a semantic coherent part and handles misalignment by adding more partitions with different granularities. The enlarged region allows the model to encode relationships between the features of more distant image areas.

Pyramid Network (Pyramid-Net)~\cite{24zheng2018coarse} tackles part misalignment by designing a pyramidal partition scheme. This scheme is similar to MGN, where the major difference is that for each of MGN's granularity, the Pyramid-Net adds one bridging part with one basic part from its adjacent parts, except for the top and bottom image areas. With this approach, some basic parts can be included in several different branches to help form coherent semantically related regions, while providing possibly richer information to the deep neural network.

The batch feature erasing (BFE) technique proposed in ~\cite{04dai2018batch} offers another way to force a deep network to learn within and between parts information. Using a batch feature erasing block in the feature erasing branch, the model training procedure implicitly asks the model to learn more robust part-level feature representations and relationships. Besides using the batch feature erasing block, using DropBlock \cite{11ghiasi2018dropblock} is also a possibility.

Most of the above mentioned methods aim to enable a deep learning model to encode local and global, within part and between parts information from the raw image. The question then becomes: \textbf{could we have a model design that enables the deep learning model to learn local and non-local information and relationships in a less hand-crafted and more data-driven way?}

In this paper, we present our perspective of incorporating non-local operations with second-order statistics in Convolutional Neural Networks (CNN) as the first attempt to model feature map correlations directly for the person re-ID problem, and propose a Second-order Non-local Attention (SONA) as an effective yet efficient module for person re-ID. By modeling the correlations of the positions in the feature map using non-local operations, the proposed module can integrate the local information captured by convolution operations into long range dependency modeling ~\cite{01wang2018non-local, 09yue2018compact, 69vaswani2017transformer}. This idea is explained in Figure~\ref{fig:figure0}. This property is appealing, since we establish a correlation between salient features captured by local convolution operations. Recent works have shown that deep convolutional networks equipped with high-order statistics can improve classification performance~\cite{19li2018towards}, and Global Second-order Pooling (GSoP) methods are used to represent the image~\cite{16lin2015bilinear, 22li2017second-order}. However, all these methods produce very high dimensional representations for the following fully connected layers, and they cannot be easily used as a building block like other first order (average/max) pooling methods. We overcome this drawback by employing the covariance matrix resulting from the non-local position-wise operations and use the matrix as an attention map.

The main contributions of our work can be summarized as follows:
\begin{itemize}
    \item{
        To overcome the well-aligned body parts limitations and to generalize part-based models, we propose a novel SONA module to model feature maps second-order correlations as an attention map directly that not only captures non-local (also local) correlations, but also the detailed salient features for person re-ID.
    }
    \item{
        To maximize the flexibility of the DropBlock mechanism and to encourage SONA to capture more distant and varied feature map correlations, we generalize DropBlock by allowing variable drop block sizes.
    }
    \item{
        In order to provide a large spatial view for the SONA module to capture more detailed spatial correlations and for the generalized version of DropBlock to further capture flexible spatial correlations, we modify the original ResNet50 using dilated convolutions.
    }
    \item{
        Our version of DropBlock and the use of the dilated convolutions complement the proposed SONA module to obtain state-of-the-art performance for person re-ID.
    }
\end{itemize}

\begin{figure*}[t]
\centering
\includegraphics[width=18cm]{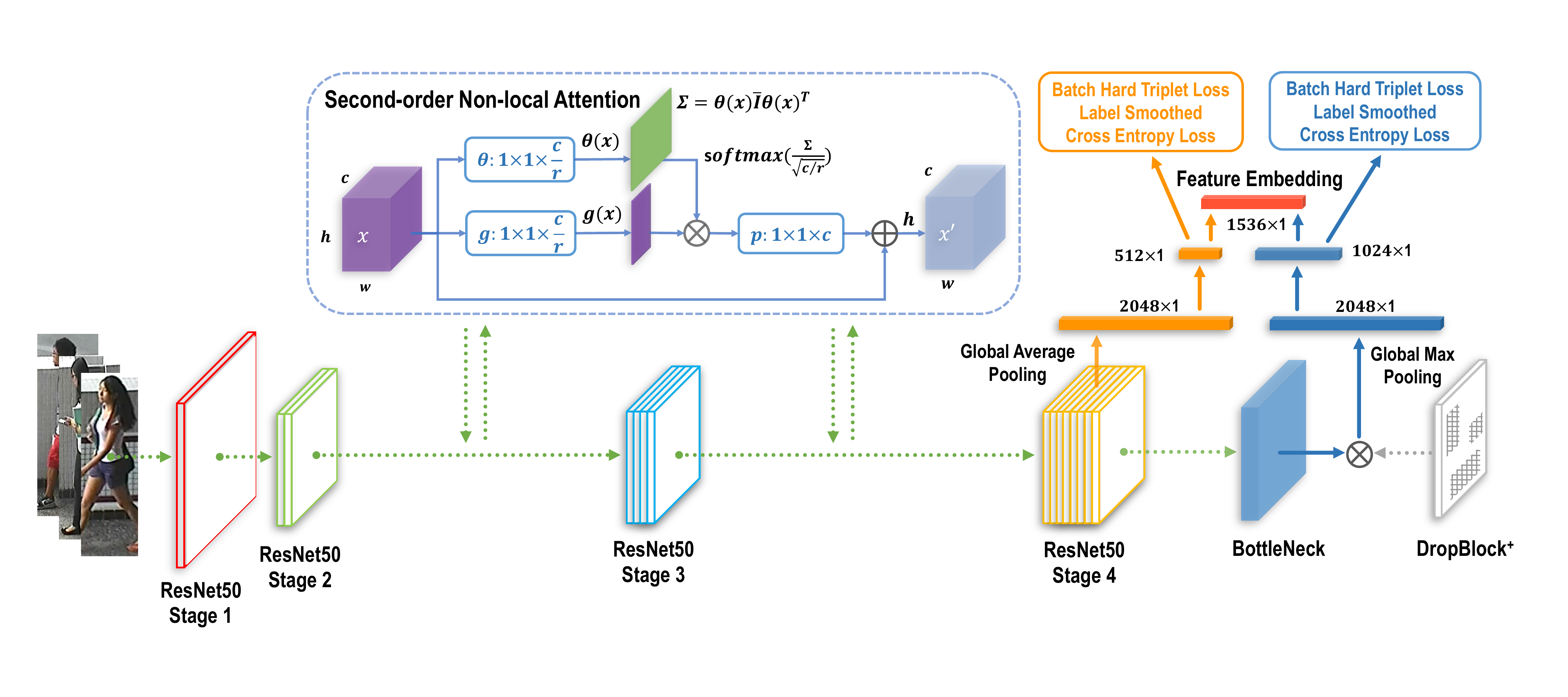}
\caption{The overall architecture of the proposed SONA-Net for the person re-ID task. The orange colored flow serves as global supervision for the blue colored feature maps region DropBlock\textsuperscript{+} branch. The SONA module can be injected after shallow stages of ResNet50. During testing, the feature embedding concatenated from both global branch and DropBlock\textsuperscript{+} is used for the final matching distance computation.}
\label{fig:nlsoa-flow-chart}
\end{figure*}

\section{Second-order Non-local Attention Network}
In this section, we describe our proposed SONA Network (SONA-Net). The network consists of (1) a backbone architecture similar to what was used in BFE~\cite{04dai2018batch}; (2) the proposed second-order non-local attention module; and (3) a generalized version of a DropBlock module, which we refer to as DropBlock\textsuperscript{+} (DB\textsuperscript{+}). The non-local attention is capable of explicitly encoding non-local location-to-location feature level relationships. DropBlock\textsuperscript{+} plays a role in encouraging the non-local module to learn more useful long distant relationships.

\subsection{Network Architecture}
\label{section:network-architecture}
Figure~\ref{fig:nlsoa-flow-chart} shows the overall network architecture, which includes a backbone network, a global branch (orange colored arrows) and a local branch (blue colored arrows), which shares a similar general architecture with BFE~\cite{04dai2018batch}.
For the backbone network, we use ResNet50~\cite{32he2016resnet} as the building foundation for feature map extraction. We further modify the original ResNet50 by adjusting the stages and removing the original fully connected layers for multi-loss training, similar to prior work~\cite{05sun2018beyond, 57li2013learning, 04dai2018batch}. In order to provide a large spatial view for the SONA module to capture more detailed spatial correlations and for the DropBlock\textsuperscript{+} to drop, we modified the original ResNet50 stage 3 and stage 4 with some dilated convolutions~\cite{yu2017dilated}, and get a larger feature map with size: $48\times16\times2048$ given the input size: $384\times128\times3$. Notice that our modified stage 3 and stage 4 share the same spatial size with the original stage 2 of ResNet50, but with doubled number of output channels.
This is particularly useful for tasks requiring localization information, such as body parts. Since each spatial position of a set of feature maps corresponds to a feature vector, and this position only provides a coarse location, while the feature vector encode more finer localization information. By keeping the same spatial size, the same position on feature map of different stages encode richer localization information when doubling the number of channels.

The global branch consists of a global average pooling (GAP) layer to produce a 2048 dimensional vector and a feature reduction module containing a 1$\times$1 convolution layer, a batch normalization layer, and a ReLU layer to reduce the dimension to 512 providing a compact global feature representation for both the triplet loss and cross entropy loss.

The local branch contains a ResNet bottleneck block~\cite{32he2016resnet}, which consists of a sequence of convolution and batch normalization layers, with a ReLU layer at the end. The feature map produced by the backbone network feeds directly into the bottleneck layer. The DropBlock\textsuperscript{+} layer modifies the DropBlock~\cite{11ghiasi2018dropblock} layer to allow a variable size for both height and width of the drop block area. We apply the mask computed by the DropBlock\textsuperscript{+} module to the feature map produced by the bottleneck block. We use global max pooling (GMP) on the masked feature map to obtain the 2048 dimensional max vector and a similar reduction module follows the GMP layer to further reduce the dimension to 1024 for both the triplet loss and cross entropy loss. The feature vectors from the global and local branches are concatenated as the final feature embedding for the person re-ID task.
As an important component of the network architecture, the SONA module is applied to the early stages of the backbone network to model the second-order statistical dependency. With the enhancement introduced by the SONA, the network is able to learn richer and more robust person identity related features.

In our work, we adopt batch hard triplet loss~\cite{39hermans2017defensetripletloss} and label-smoothed cross-entropy loss~\cite{70szegedy2016rethinking, 14xie2018bag} together to train both the global branch and local branch, respectively.

\subsection{Second-order Non-local Attention Module}
The overview of the SONA module is displayed in Figure~\ref{fig:nlsoa-flow-chart}.

Let $\mathbf{x}\in\mathbb{R}^{h\times{w}\times{c}}$ denote the input feature map for the SONA module, where $c$ is the number of channels, $h$ and $w$ are spatial height and width of the tensor. We collapse the spatial dimension into a single dimension which yields a tensor $\mathbf{x}$ with size $hw$ by $c$.

We use a 1$\times$1 convolution followed by a batch normalization layer and a Leaky Rectified Linear Unit (LeakyReLU) that forms a function called $\theta$ to reduce the number of channels $c$ to $c/r$ of the input $x$. We use a 1$\times$1 convolution that forms $g$ which serves a similar role to function $\theta$. This leads to
$\theta(\mathbf{x})$ with shape ${hw\times{\frac{c}{r}}}$ and $g(\mathbf{x})$ with shape ${hw\times{\frac{c}{r}}}$. In our experiments, we set the reduction factor $r$ to 2. The covariance matrix is computed using $\theta(\mathbf{x})$ as
\begin{equation}
    \boldsymbol{\Sigma}={\theta(\mathbf{x})}\bar{\mathbf{I}}{\theta(\mathbf{x})}^{T}
\end{equation}

where $\bar{\mathbf{I}}=\frac{1}{c/r}(\mathbf{I}-\frac{1}{c/r}\mathbf{1})$, which follows the practice in~\cite{19li2018towards}. Similar to ~\cite{69vaswani2017transformer}, we adopt $\frac{1}{\sqrt{c/r}}$ as the scaling factor for the covariance matrix before applying $softmax$, which yields
\begin{equation}
    {\mathbf{z}} = softmax(\frac{\mathbf{\Sigma}}{\sqrt{c/r}}){g(\mathbf{x})}
\end{equation}
Finally, we use a simple learnable transformation $p$, a 1$\times$1 convolution in our case, to restore the channel dimension of the attended tensor from $c/r$ to $c$, and we define the second-order non-local attention module as:

\begin{equation}
     {SONA(\mathbf{x})} = \mathbf{x} + p(\mathbf{z})
\end{equation}
 With proper reshaping, we have $SONA(\mathbf{x}) $ with shape ${h\times{w}\times{c}}$ as the input to the following ResNet50 stages as shown in Figure~\ref{fig:nlsoa-flow-chart}.

\begin{figure}[t]
\centering
\includegraphics[width=8.15cm]{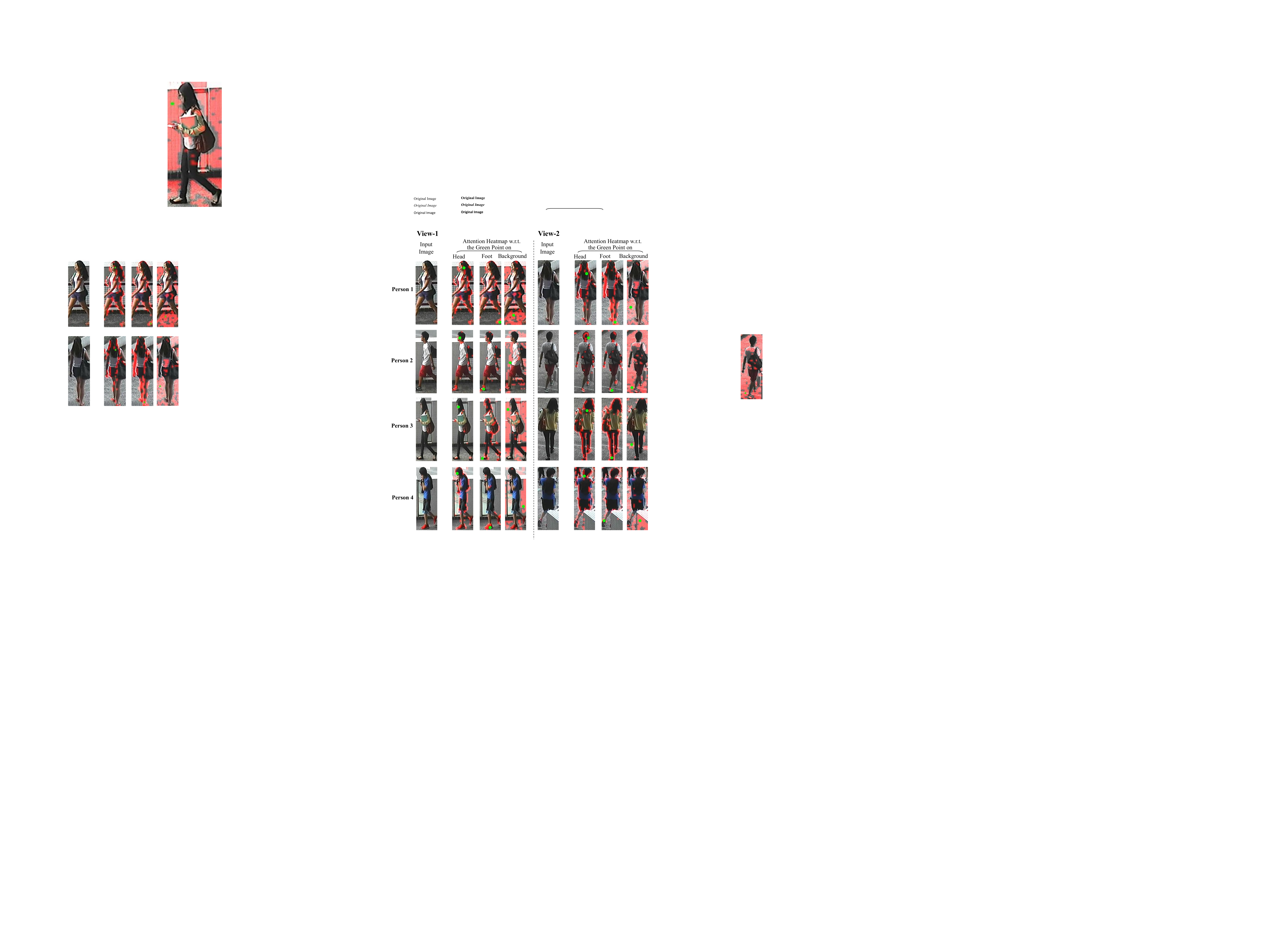}
\caption{Examples of non-local covariant attention heatmaps with different viewpoints. The green points in each heatmap are the reference points and the red points are the top related points. We can see that when the reference points (green) are located within the body region, their highly related red points are also in the body region capturing salient features such as logos on the shoes or watches. The background reference points are more related to background points.}
\label{fig:illus}
\end{figure}

We use an example to illustrate the effects of the proposed second-order non-local attention for encoding image location-to-location, human body part-to-part relationships. Given a pedestrian image $I$, assume that around image area $I(p,q)$, there is a noticeable signal (e.g., a area with high contrast), and around image area $I(p',q')$, there is another noticeable signal. After the first two/three stages of the ResNet computation, as part of the SONA module input tensor $x$, these two signals appear as features $x(p,q,:)$ and $x(p',q',:)$. The correlations between these two signals/features are then captured by computing the covariance matrix as attention for the feature tensor $x$. Using this mechanism, we explicitly tell the deep network that: (1) There are correlations between features from these two locations. (2) More attention should be spent on these locations (and their relationship) for the following computations in the deeper layers. (3) The latter layer in the deep learning model will learn under which circumstances such correlation is related (or not related) to the identity information of the person shown in the image.

We also visualize the effects in Figure~\ref{fig:illus} using different camera view images from multiple persons and the attention weights from the training process.

\begin{table}[]
\begin{scriptsize}
\begin{tabular}{|c|cclc|}
\hline
\multirow{2}{*}{Dataset} & \multirow{2}{*}{Market1501}  & \multicolumn{2}{c}{CUHK03}             & \multirow{2}{*}{DukeMTMC-reID} \\
                         &                              & \multicolumn{1}{l}{  labeled} & detected &                                \\ \hline \hline
identities               & 1501                         & \multicolumn{2}{c}{1467}               & 1812                           \\
images                   & 32668                        & 14096                       & 14097    & 36411                          \\
cameras                  & 6                            & \multicolumn{2}{c}{2}                  & 8                              \\ \hline
train IDs                & 751                          & \multicolumn{2}{c}{767}                & 702                            \\
test IDs                 & 750                          & \multicolumn{2}{c}{700}                & 1110                           \\ \hline
train images             & 12936                        & 7368                        & 7365     & 16522                          \\
query images             & 3368                         & \multicolumn{2}{c}{1400}               & 2228                           \\
gallery images           & 19732 & 5328                        & 5332     & 17661                          \\ \hline
\end{tabular}
\end{scriptsize}
\vspace{+0.21cm}
\caption{Statistics of the three evaluated re-ID datasets.}
\label{tab:datasets}
\end{table}

\section{Experimentation}
To evaluate the effectiveness of the proposed method in the person re-ID task, we perform a number of experiments using three public person re-ID datasets: Market1501~\cite{28zheng2015market1501}, CUHK03~\cite{26li2014deepreid, 27zhong2017reranking}, and DukeMTMC-reID~\cite{30zheng2017unlabeled} and compare our results with state-of-the-art methods. To investigate the effectiveness of each component and the design choices, we also perform ablation studies on the CUHK03 dataset with the new protocol~\cite{27zhong2017reranking}. Table \ref{tab:datasets} shows the statistics of each dataset.

\subsection{Datasets}
The \textbf{Market1501} dataset contains 1,501 identities collected by 5 high resolution cameras and 1 low resolution camera, where different camera viewpoints may capture the same identities. A total of 32,668 pedestrian images were produced by the Deformable Part Model (DPM) pedestrian detector. Following prior work~\cite{05sun2018beyond, 07wang2018learning, 24zheng2018coarse}, we split the dataset into training set with 12,936 images of 751 identities and testing set of 3,368 query images and 15,913 gallery images of 750 identities. Note that the original testing set contains 19,732 images, including 3,819 junk images (file names beginning with ``-1''). We ignore these junk images when matching as instructed by the dataset's website \footnote{http://www.liangzheng.org/Project/project\_reid.html}.

The \textbf{CUHK03} dataset contains manually labeled 14,096 images and DPM detected 14,097 images of a total of 1,467 identities captured by two camera views. We follow a new protocol~\cite{27zhong2017reranking} that is similar to Market1501's setting, which splits all identities into non-overlapping 767 identities for training and 700 identities for testing. The labeled dataset contains 7,368 training images, 5,328 gallery, and 1,400 query images for testing, while the detected dataset contains 7,365 images for training, 5,332 gallery, and 1,400 query images for testing.

The \textbf{DukeMTMC-reID} dataset~\cite{30zheng2017unlabeled} is a subset of the DukeMTMC dataset~\cite{29ristani2016duketmtmc}. It contains 1,404 identities captured by more than two cameras. While 408 identities only appear in one camera, they are treated as distractor identities. We follow a Market1501-like new protocol~\cite{30zheng2017unlabeled}, which splits the 1,404 identities into 702 identities with 16,522 images for training, and the other 702 identities along with those 408 distractor identities are used for testing. The testing set contains 17,661 gallery images and 2,228 query images.

\subsection{Implementation}
To capture more detailed information from each image, we resize all images to a resolution of 384$\times$128, similar to PCB. For training, we also apply the following data augmentation to the images: horizontal flip, normalization, and cutout~\cite{31devries2017cutout}. For testing we apply horizontal flip and normalization, and use the average of original feature and flipped feature for generating the final feature embedding. We use ResNet-50~\cite{31devries2017cutout}, initialized with the pre-trained weights on ImageNet~\cite{33deng2009imagenet}, as our backbone network with the modifications described above. In our variable size DropBlock layer, we set $\gamma$ to 0.1, \textit{block\_height} to 5, and \textit{block\_width} to 8. We randomly sample 32 identities, each with 4 images for the mini-batch in every training iteration. We choose Adam optimizer~\cite{34kingma2014adam} with a warm-up strategy. The initial learning rate is set to 1e-4 and increases by 1e-4 every 5 epochs for the first 50 epochs. After the warm-up, the learning rate keeps at 1e-3, then decays to 1e-4 at epoch 200, and further decays to 1e-5 at epoch 300 until a total of 400 epochs. The whole training procedure takes about 2.5 hours using 4 GTX1080Ti GPUs based on the PyTorch framework~\cite{00paszke2017automatic}. All our experimental results are reported using the same settings across all datasets.

\subsection{Comparison with State-of-the-art}
To evaluate the person re-ID performance of the proposed method and to compare the results with the state-of-the-art methods, we use cumulative matching characteristics (CMC) at \textit{Rank-1}, \textit{Rank-5}, \textit{Rank-10}, and the mean average precision (\textit{mAP}) as our evaluation metrics.

We compare our proposed method (SONA-Net) with recent state-of-the-art methods using Market1501, DukeMTMC-reID, and CUHK03. For CUHK03, we adopt the new protocol~\cite{27zhong2017reranking} similar to other methods to simplify the evaluation procedure. All reported results do \textbf{not} apply any re-ranking~\cite{27zhong2017reranking} or multi-query fusion~\cite{28zheng2015market1501} techniques. Note that most previous efforts only report the results of a single run; however, due to the randomness of the training procedure of deep neural networks, the trained model and the corresponding test performance might vary. Therefore, in order to evaluate the effectiveness of the proposed approach more fairly, we run each of our experiment configurations four times and report both the mean and standard deviation values for all four evaluation metrics. We compare our result's mean value against the existing state-of-the-art results, and mark the better result using a bold font. We use ``*'' to denote the methods that rely on auxiliary information. The compared methods can be divided into two categories according to feature types: non part-based features and part-based features. We also list the results for our model variations: SONA$^{2}$-Net, SONA$^{3}$-Net and SONA$^{2+3}$-Net, indicating the SONA module is applied after ResNet50 stage 2, stage 3, or both stage 2 and 3, and all variations share the same backbone network and DropBlock\textsuperscript{+} module.

\begin{table}
\begin{center}
\begin{tabular}{|l@{\hskip 5pt}|c@{\hskip 5pt}c@{\hskip 5pt}c@{\hskip 5pt}c@{\hskip 5pt}|}
\hline
Method	&	\textit{ mAP } 	&	\textit{Rank-1}	&	\textit{Rank-5}	&	\textit{Rank-10}	\\
\hline\hline
SOMAnet~\cite{46barbosa2018SOMAnet}	&	47.9 	&	73.9 	&	-	&	-	\\
SVDNet~\cite{43sun2017svdnet}	&	62.1 	&	82.3 	&	92.3 	&	95.2 	\\
PAN~\cite{42zheng2018pan}	&	63.4 	&	82.8 	&	-	&	-	\\
Transfer~\cite{41geng2016transfer}	&	65.5 	&	83.7 	&	-	&	-	\\
DML~\cite{40zhang2018dml}	&	68.8 	&	87.7 	&	-	&	-	\\
Triplet Loss~\cite{39hermans2017defensetripletloss}	&	69.1 	&	84.9 	&	94.2 	&	-	\\
DuATM~\cite{53si2018DuATM}   &   76.62  &    91.42   &  97.09  &  98.96   \\
Deep-CRF~\cite{10chen2018group} &   81.6    &   93.5  &   97.7    &   -   \\
BFE$^{256+512}$~\cite{04dai2018batch}   &   82.8   &    93.5   &  -  &  -   \\
BFE~\cite{04dai2018batch}   &   85.0   &    94.4   &  -  &  -   \\
\hline
MultiRegion~\cite{50ustinova2017multiregion}	&	41.2 	&	66.4 	&	85.0 	&	90.2 	\\
HydraPlus~\cite{49liu2017hydraplus}	&	-	&	76.9 	&	91.3 	&	94.5 	\\
PAR~\cite{48zhao2017PAR}	&	63.4 	&	81.0 	&	92.0 	&	94.7 	\\
PDC*~\cite{47su2017PDC}	&	63.4 	&	84.4 	&	92.7 	&	94.9 	\\
MultiLoss~\cite{45li2017multiloss}	&	64.4 	&	83.9 	&	-	&	-	\\
PartLoss~\cite{44yao2019partloss}	&	69.3 	&	88.2 	&	-	&	-	\\
MultiScale~\cite{38chen2017multiscale}	&	73.1 	&	88.9 	&	-	&	-	\\
GLAD*~\cite{37wei2017glad}	&	73.9 	&	89.9 	&	-	&	-	\\
PCB~\cite{05sun2018beyond}	&	77.4 	&	92.3 	&	97.2 	&	98.2 	\\
PCB+RPP~\cite{05sun2018beyond}	&	81.6 	&	93.8 	&	97.5 	&	98.5 	\\
MGN~\cite{07wang2018learning}	&	86.9 	&	95.7 	&	-	&	-	\\
Local-CNN*~\cite{35yang2018local}	&	87.4 	&	\textbf{95.9} 	&	-	&	-	\\
Pyramid-Net~\cite{24zheng2018coarse}    &   88.2   &    95.7   &    98.4   & 99.0   \\
\hline
SONA$^{2}$-Net $\mu$	&	88.67	&	95.68	&	98.42	&	99.03 	\\
SONA$^{2}$-Net $\sigma$	&	$\pm$0.08	&	$\pm$0.18	&	$\pm$0.08	&	$\pm$0.04 	\\
SONA$^{3}$-Net   $\mu$  &	88.63	&	95.53	&	98.48	&	99.15 	\\
SONA$^{3}$-Net $\sigma$	&		$\pm$0.08	&	$\pm$0.08	&		$\pm$0.11	&		$\pm$0.05 	\\
SONA$^{2+3}$-Net $\mu$	&	\textbf{88.83}	&	95.58	&	\textbf{98.50}	&	\textbf{99.18} 	\\
SONA$^{2+3}$-Net $\sigma$	&	$\pm$0.04	&	$\pm$0.15	&	$\pm$0.07	&	$\pm$0.13 	\\
\hline
\end{tabular}
\end{center}
\caption{Comparison of our proposed method with state-of-the-art methods for the Market-1501 dataset. $\mu$ and $\sigma$ represents mean and standard deviation of performance, respectively.}
\label{tbl:Market1501}
\end{table}
%
\textbf{Market1501.}
Table~\ref{tbl:Market1501} shows the detailed comparisons for Market1501. For this dataset, we categorize the compared methods into two groups based on the feature types, i.e., methods that explore global or local features and methods that take advantage of part information. The results show that part-based methods generally outperform methods based on global features. By integrating both global features with batch erasing local features, BFE shows competitive results compared to most part-based methods. Our approach has a similar network architecture as BFE, but BFE lacks the mechanism of modeling the information in different positions of the feature map and our model variant SONA$^{2+3}$-Net has improved the performance by 3.8\% and 1.1\% for \textit{mAP} and \textit{Rank-1} metrics, respectively. One advantage of BFE (motivating our approach) is its simplicity, while part-based methods employ complex branch settings or training procedures to coordinate the learning process of different parts. With BFE's simpler network architecture, the performance of the proposed approach is comparable to or noticeably better than the state-of-the-art part-based models, such as a newly developed Pyramid-Net.

\begin{table}
\begin{center}
\begin{tabular}{|l@{\hskip 5pt}|c@{\hskip 5pt}c@{\hskip 5pt}c@{\hskip 5pt}c@{\hskip 5pt}|}
\hline
Method	&	\textit{mAP} 	&	\textit{Rank-1}    &    \textit{Rank-5} &   \textit{Rank-10}\\
\hline\hline

SVDNet~\cite{43sun2017svdnet}	&	56.8 	&	76.7    &   86.4    &   89.9    \\
AOS~\cite{51huang2018AOS}	&	62.1 	&	79.2 	&   -   &   -   \\
HA-CNN~\cite{36li2018harmonious}	&	63.8 	&	80.5 	&   -   &   -   \\
GSRW~\cite{52shen2018GSRW}	&	66.4 	&	80.7 	&   88.5    &   90.8\\
DuATM~\cite{53si2018DuATM}	&	64.58 	&	81.82   &   90.17   &   95.38\\
Local-CNN*~\cite{35yang2018local}	&	66.04 	&	82.23 	&   -   &   -   \\
PCB+RPP~\cite{05sun2018beyond}	&	69.2 	&	83.3 	&   90.5    &   92.5\\
Deep-CRF~\cite{10chen2018group}	&  69.5 & 84.9    &   92.3    &   -   \\
BFE$^{256+512}$~\cite{04dai2018batch}   &   71.5   &    86.8 	&   -   &   -  \\
BFE~\cite{04dai2018batch}   &   75.8   &    88.7 	&   -   &   -  \\
MGN~\cite{07wang2018learning}	&	78.4 	&	88.7 	&   -   &   - 	\\
Pyramid-Net~\cite{24zheng2018coarse}	&	\textbf{79.0}	&	89.0    &   94.7    &   96.3	\\
\hline
SONA$^{2}$-Net $\mu$	&	78.05	&   	89.25	&   95.23    & 96.50  \\
SONA$^{2}$-Net $\sigma$	&	$\pm$0.38	&	$\pm$0.32 &   $\pm$0.41 &   $\pm$0.31 \\
SONA$^{3}$-Net $\mu$	&	78.18	&   	\textbf{89.55}	&   95.13    & 96.50  \\
SONA$^{3}$-Net $\sigma$	&	${\pm}$0.29	&	$\pm$0.38 &   $\pm$0.15 &   $\pm$0.22 \\
SONA$^{2+3}$-Net $\mu$	&	78.28	&   	89.38	&   \textbf{95.35}    & \textbf{96.55}  \\
SONA$^{2+3}$-Net $\sigma$	&	${\pm}$0.11 &	$\pm$0.36 &   $\pm$0.15 &   $\pm$0.11   \\
\hline
\end{tabular}
\end{center}
\caption{Comparison of the proposed method with state-of-the-art methods for the DukeMTMC-reID dataset.}
\label{tbl:DukeMTMC}
\end{table}
%

\textbf{DukeMTMC-reID.} For this dataset, Table~\ref{tbl:DukeMTMC} shows that the proposed approach achieves slightly better or comparable results compared to state-of-the-art baseline methods, such as Pyramid-Net and MGN. Similar to the comparison with Market1501, our model variants outperform both BFE and BFE$^{256+512}$, and all our model variants achieve almost the same performance. Further, the performance of the proposed method is not as sensitive to hyper-parameter settings as that of  the BFE variants, e.g., BFE and BFE$^{256+512}$ achieve 71.5\% vs. 75.8\% for \textit{mAP} and 86.8\% vs. 88.7\% for \textit{Rank-1} respectively.

\begin{table}
\begin{center}
\begin{tabular}{|l|c@{\hskip 5pt}c@{\hskip 5pt}|c@{\hskip 5pt}c@{\hskip 5pt}|}
\hline
\multirow{2}{*}{Method}	&	\multicolumn{2}{c|}{Labeled}	&			\multicolumn{2}{c|}{Detected}			\\
\cline{2-5}
	&	\textit{mAP} 	&	\textit{Rank-1}	&	\textit{mAP} 	&	\textit{Rank-1}	\\
\hline\hline
PAN~\cite{42zheng2018pan}	&	35.0 	&	36.9	&	34	&	36.3	\\
SVDNet~\cite{43sun2017svdnet}	&	37.8 	&	40.9	&	37.3	&	41.5	\\
HA-CNN~\cite{36li2018harmonious}	&	41.0 	&	44.4	&	38.6	&	41.7	\\
Local-CNN*~\cite{35yang2018local}	&	53.83 	&	58.69	&	51.55	&	56.76	\\
PCB+RPP~\cite{05sun2018beyond} &    -   &   -   &	57.5	&	63.7	\\
MGN~\cite{07wang2018learning}	&	67.4 	&	68.0 	&	66.0 	&	68.0 	\\
BFE~\cite{04dai2018batch}	&	70.9	&	75.0	&	67.9	&	72.1	\\
BFE$^{256+512}$~\cite{04dai2018batch}	&	71.2	&	75.4	&	70.8	&	74.4	\\
Pyramid-Net~\cite{24zheng2018coarse}	&	76.9	&	78.9	&	74.8	&	78.9	\\
\hline
SONA$^{2}$-Net $\mu$	&	\textbf{79.23}	&	\textbf{81.85}	&	76.35	&	79.10	\\
SONA$^{2}$-Net $\sigma$	&	$\pm$0.78	&	$\pm$0.84	&	$\pm$0.68	&	$\pm$0.56   \\
SONA$^{3}$-Net $\mu$	&	79.18	&	81.05	&	76.38	&	78.90	\\
SONA$^{3}$-Net $\sigma$	&	$\pm$0.19	&	$\pm$0.36	&	$\pm$0.88	&	$\pm$0.80	\\
SONA$^{2+3}$-Net $\mu$	&	\textbf{79.23}	&	81.40	&	\textbf{77.27}	&	\textbf{79.90}	\\
SONA$^{2+3}$-Net $\sigma$	&	$\pm$0.23	&	$\pm$0.80	&	$\pm$0.43	&	$\pm$0.67	\\
\hline
\end{tabular}
\end{center}
\caption{Comparison   of   our   proposed   method   with   state-of-the-art methods for the CUHK03 dataset using the new protocol~\cite{27zhong2017reranking}.
For the labeled set, the results of model variation SONA$^{2}$-Net at \textit{Rank-5} and \textit{Rank-10} are \textbf{92.55\%} ($\pm$0.56) and \textbf{95.58\%} ($\pm$0.61) respectively, compared to Pyramid-Net's~\cite{24zheng2018coarse} 91.0\% and 94.4\%. For the detected dataset, the results of model variant SONA$^{2+3}$-Net are \textbf{91.00\%}($\pm$0.37) and 94.48\% ($\pm$0.13), compared to Pyramid-Net's~\cite{24zheng2018coarse} 90.7\% and \textbf{94.5\%}, respectively.}\label{tbl:CUHK03}
\end{table}

\textbf{CUHK03.} This dataset is one of the most challenging person re-ID datasets due to the adoption of the new protocol with two types of person bounding boxes as described above. We can see from Table~\ref{tbl:CUHK03} that our proposed approach for the CUHK03 Labeled dataset outperforms all state-of-the-art models.
Similar to the previous comparisons with BFE and its variant on the Market1501 dataset, our proposed SONA$^{2}$-Net model variant outperforms BFE$^{256+512}$ with 8.03\% at \textit{mAP} and 6.45\% at \textit{Rank-1}, respectively. Our method achieves noticeably better performance than state-of-the-art results w.r.t. \textit{mAP}, \textit{Rank-1}, \textit{Rank-5}, and is only slightly worse than Pyramind-Net in \textit{Rank-10}. Our SONA$^{2+3}$-Net model variant exceeds BFE$^{256+512}$ 6.47\% and 5.50\% at metrics \textit{mAP} and \textit{Rank-1}.

So far, we discussed the experimental results with each dataset separately. We can also make the following general observations from these experiments:
\begin{enumerate}
    \item{
        Although we observe that for each dataset there exists a best setting, the performances of the different settings are very close to each other. In particular, even if we fix one setting randomly (or use the setting of SONA$^{2}$-Net, which has the fewest parameters), we can still outperform the baselines for most metrics. The stability of the second-order non-local attention module in different settings makes it flexible and easy to be applied to another different network architecture without the need for additional hyper-parameter tuning.
    }
    \item{
        Our approach achieves consistent improvements for the four datasets. Nevertheless, we find that we obtain most improvement for CUHK03 (2.33\%), while we see the smallest improvement for Market1501 (0.63\%) at \textit{mAP} compared with the closest known model. This is understandable, because the different characteristics of the datasets, such as the used bounding box detection algorithm and misalignment of different parts rooted in the model design. Most previous approaches also have a larger performance variance on different datasets, e.g., MGN performs much worse with the CUHK03 Labeled dataset than with the Market1501 dataset (11.83\% and 1.93\% worse than the proposed approach for \textit{mAP} , respectively), which is probably due to its part-based mechanism being sensitive to the accuracy of the bounding box detection and accurate part alignment. Pyramid-Net mitigates this problem by sharing a common basic part between adjacent parts.
    }
\end{enumerate}

\begin{table*}[ht]
\centering
\footnotesize
\begin{center}
\begin{tabular}{|l@{\hskip 4pt}|c@{\hskip 4pt}c@{\hskip 4pt}c@{\hskip 4pt}c@{\hskip 4pt}|c@{\hskip 4pt}c@{\hskip 4pt}c@{\hskip 4pt}c@{\hskip 4pt}|c@{\hskip 4pt}c@{\hskip 4pt}c@{\hskip 4pt}c@{\hskip 4pt}|c@{\hskip 4pt}c@{\hskip 4pt}c@{\hskip 4pt}c@{\hskip 4pt}|}
\hline
\multirow{2}{*}{Models} & \multicolumn{4}{c@{\hskip 4pt}|}{CUHK03 Labeled} & \multicolumn{4}{c@{\hskip 4pt}|}{CUHK03 Detected} & \multicolumn{4}{c@{\hskip 4pt}|}{DukeMTMC-reID} & \multicolumn{4}{c@{\hskip 4pt}|}{Market-1501} \\ \cline{2-17}
                        & mAP   & R-1    & R-5    & R-10   & mAP    & R-1     & R-5    & R-10   & mAP  & R-1   & R-5   & R-10 & mAP  & R-1  & R-5  & R-10   \\ \hline \hline
MGN~\cite{07wang2018learning}	&	67.4 	&	68.0    &   -   &   -	&	66.0 	&	68.0    &   -   &   -	&   78.4    &   88.7    &    -   &   -   &   86.9    &   \textbf{95.7}    &   - &   -\\
BFE~\cite{04dai2018batch}   & 70.9   &   75.0    &   -   & -   &   67.9    &   72.1   &    -   &   -   &    75.8 &  88.7    &   -   & -    &   85.0   &    94.4   &  -  &  -  \\
BFE$^{256+512}$~\cite{04dai2018batch}   & 71.2   &   75.4    &   -   & -   &   70.8    &   74.4   &    -   &   -   &    71.5 &  86.8    &   -   & -    &   82.8   &    93.5   &  -  &  -  \\
Pyramid-Net~\cite{24zheng2018coarse}            & 76.9  & 78.9  & 91.0 & 94.4 & 74.8 & 78.9 & 90.7   & \textbf{94.5}   & \textbf{79.0}    & 89.0   & 94.7   &    96.3     & 88.2   & \textbf{95.7}   & 98.4   & 99.0  \\

\hline

BL $\mu$                      & 77.05  & 79.70  & 91.52 & 94.98 & 74.00 & 76.70 & 89.45   & 93.45 & 76.2    & 88.00   & 94.60   & 96.20    & 87.50   & 95.18   & 98.28   & 99.03\\
BL $\sigma$                                    & $\pm$0.22   & $\pm$0.44   & $\pm$0.33  & $\pm$0.19  & $\pm$0.30  & $\pm$0.30  & $\pm$0.25    & $\pm$0.55  & $\pm$0.30    & $\pm$0.60    & $\pm$0.10    & $\pm$0.10   & $\pm$0.16   & $\pm$0.19   & $\pm$0.04   & $\pm$0.04  \\

BL+DB $\mu$              & 77.02  & 79.13   &   90.90   &   94.88   & 74.50 & 77.25   &   89.38   &   93.03   & 76.93    & 88.28  &   94.65   &   96.15    & 87.60   & 95.43   & 98.25 &   99.03\\
BL+DB $\sigma$            & $\pm$0.28  & $\pm$0.42  & $\pm$0.21 & $\pm$0.15   & $\pm$0.37 & $\pm$0.57   & $\pm$0.27 & $\pm$0.30 & $\pm$0.24    & $\pm$0.23  & $\pm$0.11   & $\pm$0.18    & $\pm$0.07   & $\pm$0.15 & $\pm$0.15 & $\pm$0.04   \\

BL+DB\textsuperscript{+} $\mu$              & 77.45  & 79.10  & 91.60 & 94.78 & 74.30 & 76.93 & 89.50   & 93.20    & 76.95    & 88.60   & 94.88   & 96.25   & 87.68   & 95.18   & 98.32   & 99.0\\
BL+DB\textsuperscript{+} $\sigma$            & $\pm$0.54  & $\pm$0.78  & $\pm$0.31 & $\pm$0.29 & $\pm$0.25 & $\pm$0.50 & $\pm$0.25   & $\pm$0.16   & $\pm$0.15    & $\pm$0.39   & $\pm$0.18   & $\pm$0.15  & $\pm$0.16   & $\pm$0.15   & $\pm$0.08   & $\pm$0.07 \\

BL+BFE $\mu$                      & 77.20  & 79.83  & 91.03  & 94.90  & 74.85 & 77.48  & 89.95  & 93.48 & 76.85    & 88.15  & 94.6  & 95.98   & 87.73   & 95.30  & 98.35  & 99.00   \\
BL+BFE $\sigma$                                    & $\pm$0.12   & $\pm$0.47   & $\pm$0.08   & $\pm$0.14   & $\pm$0.44  & $\pm$0.58   & $\pm$0.09   & $\pm$0.26  & $\pm$0.34    & $\pm$0.50   & $\pm$0.32   & $\pm$0.11 & $\pm$0.04   & $\pm$0.29   & $\pm$0.11   & $\pm$0.10 \\

BL+SONA$^{2}$ $\mu$               & 78.48  & 80.78  & 92.03  & 95.50  & 76.20 & 78.93  & 90.40  & 94.48 & 78.18    & 89.55  & 95.05  & 96.45   & 88.50   & 95.58  & 98.32  & 99.00   \\
BL+SONA$^{2}$ $\sigma$              & $\pm$0.33  & $\pm$0.13  & $\pm$0.29  & $\pm$0.41 & $\pm$0.32 & $\pm$0.41  & $\pm$0.36  & $\pm$0.50 & $\pm$0.15    & $\pm$0.32  & $\pm$0.32  & $\pm$0.21    & $\pm$0.12   & $\pm$0.19  & $\pm$0.08  & $\pm$0.12   \\

BL+BFE+SONA$^{2}$ $\mu$                & 79.15  & 81.68  & 92.25  & 95.38  & 76.00 & 78.83  & 90.23  & 94.10 & 77.98    & 88.90  & 95.05  & 96.25   & 88.63   & 95.60  & 98.30  & 99.00 \\
BL+BFE+SONA$^{2}$ $\sigma$               & $\pm$0.11  & $\pm$0.35  & $\pm$0.09  & $\pm$0.11    & $\pm$0.43 & $\pm$0.66  & $\pm$0.18  & $\pm$0.21 & $\pm$0.04    & $\pm$0.16  & $\pm$0.11  & $\pm$0.11   & $\pm$0.04   & $\pm$0.39  & $\pm$0.16  & $\pm$0.07   \\

\hline

SONA$^{2}$-Net $\mu$               & \textbf{79.23}  & \textbf{81.85}  & \textbf{92.55} & \textbf{95.58} & 76.35 & 79.10 & 90.25   & 94.03   & 78.05    & 89.25   & 95.23   & 96.50   & 88.67   & 95.68   & 98.42   & 99.03 \\
SONA$^{2}$-Net $\sigma$              & $\pm$0.78  & $\pm$0.84  & $\pm$0.56 & $\pm$0.61 & $\pm$0.68 & $\pm$0.56 & $\pm$0.53   & $\pm$0.55 & $\pm$0.38    & $\pm$0.32   & $\pm$0.41   & $\pm$0.31     & $\pm$0.08   & $\pm$0.18   & $\pm$0.08   & $\pm$0.04  \\
SONA$^{3}$-Net $\mu$                  & 79.18  & 81.05  & 92.10 & 95.45 & 76.38 & 78.90 & 90.68   & 94.35   & 78.18    & \textbf{89.55}   & 95.13   & 96.50     & 88.63   & 95.53   & 98.48   & 99.15\\
SONA$^{3}$-Net $\sigma$                   & $\pm$0.19  & $\pm$0.36  & $\pm$0.33 & $\pm$0.45 & $\pm$0.88 & $\pm$0.80 & $\pm$0.53   & $\pm$0.45  & $\pm$0.29    & $\pm$0.38   & $\pm$0.15   & $\pm$0.22    & $\pm$0.08   & $\pm$0.08   & $\pm$0.11   & $\pm$0.05  \\
SONA$^{2+3}$-Net $\mu$              & \textbf{79.23}  & 81.40  & 92.35 & 95.57 & \textbf{77.27} & \textbf{79.90} & \textbf{91.00}   &  94.48    & 78.28    & 89.38   & \textbf{95.35}   & \textbf{96.55}   & \textbf{88.83}   & 95.58   & \textbf{98.50}   & \textbf{99.18}  \\
SONA$^{2+3}$-Net $\sigma$               & $\pm$0.23  & $\pm$0.80  & $\pm$0.09 & $\pm$0.04 & $\pm$0.43 & $\pm$0.67 & $\pm$0.37 & $\pm$0.13   & $\pm$0.11    & $\pm$0.36   & $\pm$0.15   & $\pm$0.11   & $\pm$0.04   & $\pm$0.15  & $\pm$0.07   & $\pm$0.13  \\ \hline
\end{tabular}
\end{center}
\caption{Comparison of the proposed model and its variants with MGN, BFEs, and Pyramid-Net. ``BL'' represents the Baseline network with backbone, global branch, and local branch. ``DB'' represents the original DropBlock module, and ``DB\textsuperscript{+}'' represents the DropBlock\textsuperscript{+} module. The ``SONA$^{\{2, 3, 2+3\}}$-Net'' represents the network with all components (BL, DB\textsuperscript{+}, and SONA injection variants).}
\label{tbl:ablation}
\end{table*}

\subsection{Ablation Studies}
To further investigate each component's contribution to the whole network, we perform ablation studies by deliberately removing certain modules and comparing the results for all four metrics. The overall settings remain exactly the same, while only the module under investigation is added or removed from the whole network. Specifically, in Table~\ref{tbl:ablation}, the Baseline network is the network with backbone, global branch, and local branch. Note that the Baseline network is also an improved architecture based on BFE as discussed in Section~\ref{section:network-architecture}. The DropBlock\textsuperscript{+} represents the variant with both Baseline network and the DropBlock\textsuperscript{+} module. The SONA-Net variants contain both Baseline network and DropBlock\textsuperscript{+} module. As shown in Table~\ref{tbl:ablation}, we observe that:

\begin{enumerate}
    \item{
        The Baseline network has a simple two branch architecture, but it is very effective, indicating that our architecture modification to BFE is useful. On the CUHK03 Labeled dataset, it even slightly outperforms the Pyramid-Net. When adding DropBlock\textsuperscript{+} to the Baseline network, it can improve the Baseline network in general. For other datasets, our Baseline network achieves comparable results to Pyramid-Net; only the \textit{mAP} on DukeMTMC-reID is worse than Pyramid-Net.
    }

    \item{
        When the proposed second-order non-local attention module is added in addition to the DropBlock~\textsuperscript{+} module, the overall deep network can further achieve noticeably better results than the state-of-the-art Pyramid-Net. However, different datasets have their own characteristics, and the SONA module works slightly different on those datasets. But in general, all three SONA model variants achieve similar results.
    }

    \item{
        We further conduct experiments to see if the proposed second-order non-local model works in deeper positions of the DNNs. Specifically, we place the SONA module right after Stage-4 on the global branch and found that the performance drops greatly, e.g., 75.8\% at \textit{mAP}, and 78.9\% at \textit{Rank-1} on CUHK03 Labeled dataset. We also observe a similar behavior for placing the SONA after Stage-4 on the local branch and on both branches. This indicates that the proposed SONA module, although it shows stability when placed in different earlier stages, is not appropriate for placement in later stages. This is because the purpose of the second-order non-local attention module is to capture the non-local correlation in early stages, which contains more fine-grained information.
    }

    \item{
        SONA, whenever it is applied to a model, always leads to a significant performance gain. DropBlock\textsuperscript{+} as a generalized version of DropBlock further enhances the flexibility of our proposed model. When applied together with SONA, we show that DropBlock\textsuperscript{+} yields slightly better results than BFE. Overall, DropBlock, BFE, and  DropBlock\textsuperscript{+} serve very similar purposes as regularization. We show in the experiments that they do not yield major performance improvements to the overall system. The major performance gain is from the use of our proposed SONA.
    }

    \item{
        In addition to the improvement of the test performance, we also find that the training losses are also affected by different modules. Initially, while the Baseline network is not affected by other modules, it produces relatively small loss. However, when we add the DropBlock\textsuperscript{+} to the Baseline network, the average loss increases by 0.45\%. This is as expected, because DropBlock\textsuperscript{+} is essentially a regularization method preventing the network from overfitting. We further add the SONA module after the second stage and the third stage and the average losses are then 0.02\% and 0.06\% lower than the Baseline loss. This behavior indicates that the SONA module helps the training.
    }
\end{enumerate}

Overall, we demonstrate the effectiveness of our proposed second-order non-local attention for encoding non-local body part relations for person re-ID tasks.

\subsection{Inference Time Cost}
We measure the single image inference time (with ten runs) using one Nvidia Titan Xp and Market1501. The time cost for one forward pass on the model with the SONA$^{2}$ module is 8.44 ms $\pm 0.09$ ms, and 7.89 ms $\pm$0.16 ms without the SONA$^{2}$ module. The results show that the overhead caused by our SONA module is negligible.

\section{Conclusions}
In this paper, we present a new perspective of modeling feature map correlations using second-order statistics and design an attention module based on this correlation for person re-identification. By design, our model is able to capture the correlations between salient features from any spatial locations of the feature map in the early stages. Therefore, it does not rely on special part partition schemes or arrangements to handle part misalignment issues. It provides a more general, automatic, and advanced data modeling scheme for the deep neural network to learn more discriminative and robust representations in the person re-ID task. With the help of the proposed attention module, our model pushes the state-of-the-art further and achieves better results on three popular person re-ID datasets. Specifically, on the CUHK03 dataset, our model outperforms the currently best model by a noticeable margin under the new protocol. Note that we ran four experiments for the same network configuration and reported the mean and standard deviations for all four evaluation metrics in addition to the single-query and re-ranking free evaluation protocol.

{\small
\bibliographystyle{ieee_fullname}
\bibliography{egbib}
}

\end{document}